\documentclass{article}
\pdfoutput=1
\usepackage{arxiv}

\usepackage[utf8]{inputenc} 
\usepackage[T1]{fontenc}    
\usepackage{hyperref}       
\usepackage{url}            
\usepackage{booktabs}       
\usepackage{amsfonts}       
\usepackage{amsmath}
\usepackage{tabularx}
\usepackage{makecell}
\usepackage{graphicx}
\usepackage{doi}
\usepackage{tikz}
\usepackage{tikz-3dplot}
\usetikzlibrary{shapes.geometric, arrows, positioning, calc, patterns, 3d}

\tikzstyle{nn-data} = [rectangle, rounded corners, align=center, text centered, draw=black]
\tikzstyle{nn-layer} = [rectangle, text centered, align=center, draw=black]

\graphicspath{ {./images/} }

\title{Inductive biases and Self Supervised Learning in modelling a physical heating system}

\renewcommand{\headeright}{}
\renewcommand{\undertitle}{}

\date{April 2021}	

\author{Cristian Vicas\\
	Computer Science Department\\
	Tehnical University of Cluj-Napoca\\
	Romania\\
	\texttt{cristian.vicas@cs.utcluj.ro} \\
}
\hypersetup{
pdftitle={Inductive biases and Self Supervised Learning in modelling a physical heating system},
pdfsubject={cs.LG, ACM I.2.8},
pdfauthor={Cristian Vicas},
pdfkeywords={deep learning, filters, self-supervised-learning, model-predictive-control},
}

\begin{document}
\maketitle

\begin{abstract}
Model Predictive Controllers (MPC) require a good model for the controlled process. In this paper I infer inductive biases about a physical system. I use these biases to derive a new neural network architecture that can model this real system that has noise and inertia. The main inductive biases exploited here are: the delayed impact of some inputs on the system and the separability between the temporal component and how the inputs interact to produce the output of a system.

The inputs are independently delayed using shifted convolutional kernels. Feature interactions are modelled using a fully connected network that does not have access to temporal information.

The available data and the problem setup allow the usage of Self Supervised Learning in order to train the models. The baseline architecture is an \textit{Attention} based Reccurent network adapted to work with MPC like inputs. The proposed networks are faster, better at exploiting larger data volumes and are almost as good as baseline networks in terms of prediction performance. The proposed architecture family called \textit{Delay} can be used in a real scenario to control systems with delayed responses with respect to its controls or inputs. Ablation studies show that the presence of delay kernels are vital to obtain any learning in proposed architecture.

Code and some experimental data are available online.

\end{abstract}


\section{Introduction}

Model Predictive Control (MPC) is a branch of Data Driven Control that optimizes a process mainly by using a mathematical model of the system. While MPC also relies on feedback, a good system model can dramatically increase the optimization abilities of the MPC \cite{brunton_kutz_2019}. The model in a MPC loop must output the future values of a target given a set of possible commands. The model has direct or indirect access to observed past data, including past commands, target values, system sensors, etc \cite{quin_mpc_2000}. Current paper focuses only on the modelling part of the MPC loop (System Identification). 

In Data Driven Control, if the system responds fast with respect to its controls then Extermum Seeking Control can be employed \cite{bruton_extremum_2010}. MPCs are more fitted where the feedback cannot be directly used to control the system.

For almost any machine learning (ML) problem, some function aproximator must determine a function $f$, such that, given a set of observations $x$ and labels $y$ the following holds: $f(x) \approx y$.  Here the labels are derived directly from the data. This frames the problem in the realm of Self Supervised Learning \cite{Goyal2021}. Another engineering area used here is the Differentiable Programming paradigm \cite{baydin_auto_diff_2017}. The focus is on the "forward" step of the method: the actual parameter values will be found using gradient descent and automatic differentiation. There are some restrictions on the differentiability of the forward steps and one has to consider the dynamic ranges of the parameters. Self Supervised Learning and Differentiable Programming are the two main paradigms that inspired this paper.  

ML and especially Neural Networks (NN) can work directly with raw inputs (no feature engineering needed). System states can be inferred from past observations (no need to formulate or to keep track of the current system state) and most importantly, a NN is fully differentiable. One can take the MPC objective and compute the "best" commands to minimize/maximize that objective using the same gradient based tools that trained the network. Of course, there is no guarantee that this new problem space is convex. 

There are a plethora of pretrained NN models in vision, sound, text. These models can be exploited to create subnetworks that can ingest images, sound and produce a "feature" for the upstream network. This mix-and-match tehnique is well known and used in the NN industry (taking a pretrained body and fine-tune the head of the network for a particular task). These abilities (pretrained models and fine-tuning) can allow one to integrate multimodality sensors and train an end-to-end model without the need of gigantic data volumes.
 
Present paper can be framed in many ways. Combination of sensors inside and outside the system, all collaborating to perform a decision is similar to how the propioceptive system works in mamals. Getting feedback from the environment and modelling the world are specific to Reinforcement Learning. While there might be promising algorithms in these fields, the paper will not focus on them. Keen reader can benchmark those algorithms against the models presented here. A curated data sample and the code is publicly available.

A MPC control loop needs to inject into the system model the state of the system and a series of potential commands. The outputs of the model will be evaluated with respect to some objective and the first entry of the best command set will be applied to the real system. The model must be able to ingest both the state of the system and future commands. These two entities have different meaning and must be handled separately. Current paper is a continuation of \cite{vicas_iccp2020} where several popular general purpose neural networks were adapted to this particular problem setup (two different inputs and one output). In present paper I exploit the known \textit{a priori} facts about the modelled system to deduce some intuitions and I try to use these intuitions as inductive biases in the NN architecture design. The goal is to get smaller, faster and/or better models than the previously developed ones. Because of this, the baseline for current experiments are the best results obtained in \cite{vicas_iccp2020}. 

First inductive bias that is introduced here is that some past value can affect the next system output. The second bias is that one can decouple finding these delays with the problem of finding the transfer function of the system without delays. Modelling the delays and the separability are the key elements exploited in the architectures proposed here. 

The delays are modelled here by convolving the signal with offseted kernels \cite{vicas_curvilinear_2015}. Great care is taken designing the filters so their parameters take values easily learnable by a neural network. The non delayed system identification is left for a fully connected network that will infer the output by combining the input features with the delay removed at the previous step. A practicioner can replace this general FCN with another differentiable architecture that exploits known feature interactions. 

The term bias is used here with the meaning "error". The assumption that the signals have a delay and this delay is constant for a problem, is of course, false. However, it approximates the reality good enough so the advantages that this assumption brings to the solution have greater utility than the errors it is generating. The usefulness can be in terms of speed, model size, accesibility, etc. From a good inductive bias one can expect some drop in prediction performance with massive gains in other areas.

The studied problem of controlling a household heating system might look trivial but there are critical areas where even a fraction of degree variation can tremendously impact the performance of the system. Some examples: Voltage references in instrumentation (eg high energy physics) where a temperature shift biases the result; Recording studios, where there is a lot of human activity in a closed space and where temperature/humidity oscillations due to HVAC system detune the musical instruments, etc.

A literature survey shows that there are many applications of using Machine Learning in Optimal Control field. System identification with reccurent neural networks is performed in \cite{2017_lstm_dynamic_identif_wang}; deriving PID controller parameters is done in \cite{2017_pid_ml_africon}. The authors extend a classical PID controller with a specialized form of neural networks in \cite{Gunther_2020, Genc2017}, or using Evolutionary programming paradigm \cite{Nyberg2017OptimizingPP}. In \cite{li2021fourier} complex physical problems are being modelled with a fully learnable set of filters, the bias that is introduced is that low frequency signals affect the future evolution of the system in a larger way than higher frequency signals/changes. Most of the papers employing a ML as the model in an MPC loop remain on the realm of classical, supervised learning framework \cite{sensors_greenhouse, israel_ml_mpc}.  A more comprehensive survey was performed in \cite{vicas_iccp2020}.

A neural network that accepts both current state and future actions is used in reinforcement learning. However, to my knowledge, using this type of input to perform system modelling for a MPC loop, is not yet a widely used idea \cite{learning_mpc_zurich}. However, the ML field is in hyperinflation mode and papers might be available.

This paper has the following contributions:
\begin{itemize}
	\item It derives a NN architecture based on two strong inductive biases: modelling the delay of the input signal and the separability between temporal interactions and feature interactions inside the system.
	\item The delay is modelled using four types of convolution filters that are adapted to work into a NN setup.
	\item Extensive experiments show that this new architecture can reach close performance to the existing "SOTA" even if they are faster and smaller.
	\item Code and a sample of the real data is available\footnote{ https://github.com/cristi-zz/auto{\_}iccp2020}.
\end{itemize}

\section{Proposed architecture}
\subsection{Overview}

The powerful reccurent architectures (GRU, LSTM) are notoriously hard to paralelize \cite{lstm_Schmidhuber, Schmidhuber90makingthe}. In recent years a strong trend emerged to replace these reccurent architectures with something that is easier to paralellize \cite{vaswani_attention}; hence the current focus, in literature, on \textit{Transformers}. Transformer-like architectures drop the temporal dependency with the cost of increased data requirements and computing power. Here, I exploit a different avenue: I use domain knowledge to drive the architecture. Networks that are not recurrent train faster using massively parallel architectures (GP-GPU) and are easier to be implemented in real time scenarios (eg FPGA or custom ASICs).

The heat moves into a system from a hot region to a cold region, with a "speed" depending on the temperature gradient. In empirical terms, the temperature in a region will change with a delay depending on the distance, temperature gradient and quality of materials. In the \textit{Delay} architecture the temperature gradient dependency is approximated (Hence naming these choices biases). The network can learn several delays that are data independent. In principle, it is straigtforward \cite{nips_spatial_transform} to design the filtering layers in such a way that their parameters are learned from the input data, with the cost of increased complexity. Ignoring the temperature gradient dependency was a deliberate decision to reduce the network complexity and capacity.

Framing into signal processing terms, (the signal noted at the source) will influence the measuring point: attenuated and with a different phase. The role of the delay layers presented here is to model this delay+attenuation, letting the accros-features block to deal only with feature-feature interaction without the time component.

In short, the architecture is framed in Self Supervized Learning. Introduced biases help reduce the network complexity.

\subsection{Data and system}

The data format and the practical application are similar to those in \cite{vicas_iccp2020}

I experiment on data from a household heating system composed from a central heater, high inertia radiating elements and a bunch of sensors mostly monitoring temperatures in and around the system.

There are 14 values collected at each time step. Five of them are very important (air/heating fluid temperatures and commands given to actuators) and the other 9 give hints about the system (tap water temperature, status of various other pumps in the system, etc). Most experiments presented here were performed on the set of 5 features.

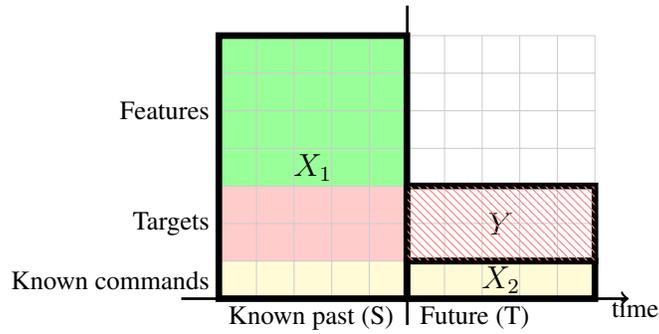
\begin{figure}[hp]
	\begin{center}
		\begin{tikzpicture}[scale=0.5]
			\fill [fill=yellow!20!white] (0, 0) rectangle (10,1);  
			\fill [fill=red!20!white] (0, 1) rectangle (5,3);    
			\fill [fill=green!40!white] (0, 3) rectangle (5,7);  
			\node [right] at (5.1, -0.5) {Future (T)};
			\node [right] at (10.2, -0.25) {time};
			\node [left] at (4.9, -0.5) {Known past (S)};
			\draw [very thin, color=black!20] (0,0) grid (10, 7);
			\draw [line width=1] (5.01, -0.7) -- (5.01, 7.8); 
			\draw [line width=1, ->] (-1, 0) -- (10.9, 0);      
			\draw [line width=2.5] (0,0) rectangle (5, 7) node[pos=0.5, font=\large] {$X_1$};
			\draw [line width=2.5] (5,0) rectangle (10, 1) node[pos=0.5, font=\large] {$X_2$};
			\draw [line width=2.5] (5,1) rectangle (10, 3) node[pos=0.5, font=\large] {$Y$};
			\draw [pattern color=red!50, pattern=north west lines] (5, 1) rectangle (10, 3);    
			\node [left] at (0, 0.5) {Known commands};
			\node [left] at (0, 2) {Targets};
			\node [left] at (0, 5) {Features};
		\end{tikzpicture}
		\caption{Data division for a training sample}
		\label{fig:data_division}
	\end{center}
\end{figure}

Figure \ref{fig:data_division} shows how a sample of data is split into inputs ($X_1$, $X_2$) and targets ($Y$) for the ML system. 

\subsection{Filters and other custom blocks}

The first inductive bias that is introduced here is the assumption that, for predicting an output value at time $t$, information from the past is relevant but with a certain latency. Each feature can have a different delay and there might be the need for more delay values from a specific feature. The filtering layers act only along-time axis. Each filter type can have side effects depending on its parameters: smoothing (low pass filtering), bandpass filtering, etc.

The input signal for a particular feature is convoluted with the filter. The result has the same time length as the input. After the convolution is applied, a batch normalization operation is applied on the result.

The filters are grouped in a filter bank. The parameters of this filter bank layer are:

\begin{itemize}
	\item Filter type: Identity, Gauss, Affine, Gabor, LogNormal, see below;
	\item The number of filters (greater than 0) per input feature. Each filter is learned and applied independently.
	\item Mode of the filter. (If the filters will be applied one per feature or one per output cell)
\end{itemize}

Regularly, one would want that each input feature to have its own learned filter. This is the \textit{one-filter-per-feature} mode. For 5 features in the input and two filters per feature, there will be 5 x 2 learnable filters.  In some special cases it might be adivsable to have a time-dependent component. For this purpose, in present paper, one can have a filter (or a filter set) for each cell in the output. Each possible output has a filter that "sees" the whole time segment of that particular feature. The convolution result has the size of 1. This mode has the advantage of decoupling the input time length of the output time length while letting each output to be affected by the the full time aperture.

In \textit{one-filter-per-cell} mode, for an output of 5 features and 10 time steps, with two filters per feature, the filter bank will have 5 x 2 x 10 independent filters. As for single kernel per feature setup, a batch normalization is applied for each cell in the output.

\begin{figure}[htbp]    
	\begin{center}
		\begin{tikzpicture}[scale=0.6, x={(0.7cm,-0.5cm)}, y={(1cm,0cm)}, z={(0cm,1cm)},] 
			
			\colorlet{kernel1}{red!70!black}
			\colorlet{kernel2}{green!70!black}
			\colorlet{kernel3}{blue!70!black}
			
			\begin{scope}[plane origin={(0, 0, 0)}, plane x={(1, 0, 0)}, plane y={(0, 1, 0)}, canvas is plane]  
				\fill [color=kernel1!15!white] (0,0) rectangle (1,12);
				\fill [color=kernel2!15!white] (1,0) rectangle (2,12);
				\fill [color=kernel3!15!white] (2,0) rectangle (3,12);
				\draw [->, thick] (0, 12) -- (3.5, 12) node[below] {features};  
				\draw [->, thick] (0, 0) -- (0, 12.5) node [above, ] {time};   
				\draw [step=1] (0, 0) grid (3, 12);
			\end{scope}

			\begin{scope}[plane origin={(0.5, 5.5, 1.5)}, plane x={(0.5, 6.5, 1.5)}, plane y={(0.5, 5.5, 2.5)}, canvas is plane]   %
				\draw[->] (0,0) -- (0, 1);
				\draw (-1,0) -- (1, 0);
				\draw [color=kernel1, thick, domain=-5:5, samples=80] plot({\x} ,{1 * exp(-(\x-2.3)^2/0.1)});		
			\end{scope}	
			
			\begin{scope}[plane origin={(1.5, 5.5, 1.5)}, plane x={(1.5, 6.5, 1.5)}, plane y={(1.5, 5.5, 2.5)}, canvas is plane]   %
				\draw[->] (0,0) -- (0, 1);
				\draw (-1,0) -- (1, 0);
				\draw [color=kernel2, thick, domain=-5:5, samples=40] plot({\x} ,{1 * exp(-(\x-3.3)^2/2.4)});		
			\end{scope}	
			
			\begin{scope}[plane origin={(2.5, 5.5, 1.5)}, plane x={(2.5, 6.5, 1.5)}, plane y={(2.5, 5.5, 2.5)}, canvas is plane]   %
				\draw[->] (0,0) -- (0, 1);
				\draw (-1,0) -- (1, 0);
				\draw [color=kernel3, thick, domain=-5:5, samples=40] plot({\x} ,{1 * exp(-(\x - 1.1)^2/0.7)});		
			\end{scope}	
		\end{tikzpicture}
		\caption{Bank filter operation for \textit{one-filter-per-feature} setup. The data consists of 3 features and 12 time steps. For each feature there is one filter in the filter bank. The convolution is performed in such a way that the output has the same length as the input. Setting the number of filters to a value greater than 1 will multiply the number of features in the output. The small vertical arrows show the centerpoint of each filter. Note that each filter have a different offset.}
		\label{fig:filter_banks}
	\end{center}
\end{figure}
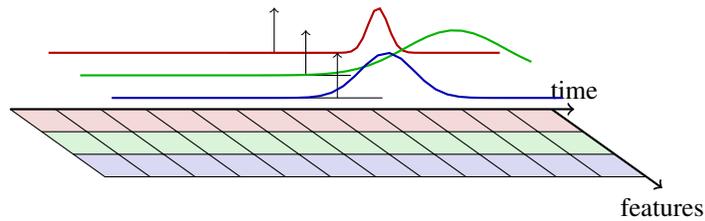

In Figure \ref{fig:filter_banks} one can see how the filter bank works for a \textit{one-filter-per-feature} setup and some Gaussian filters. All filters have been tested on simple problems to verify that there is learning happening and they behave as expected. I took extra care in making sure that the distribution of the learned parameters and the outputs follow a zero centered distribution. This is extremly important for DL architectures because the learning happens around zero. Moreover, large activations inside the network lead to vanishing/exploding gradient issues. Large parameter values lead to a slow learning and possible convergence issues. Also, the filters were altered in such a way that a parameter value close to 0 leads to a filter that mostly passes the input to the output.

Note that there is another bias introduced here, the fact that there is a fixed number of ways in which an input signal can affect the output (w.r.t. to delay). This number [the number of filters in filtering block] is a hyperparameter for the architecture. 

Below I describe each filter that can pe instantiated in the filter bank.

\subsubsection{Identity transform}

This filter lets the input pass to the output unchanged. If the output time dimension is smaller than the input time dimension then the past data is cropped out from the input. There is no Batch Normalization (BatchNorm) \cite{batch_ioffe2015} or any other transformation applied to the data.

\subsubsection{Affine transform}

The 1D signal is scaled with a value $a=e^{5s}$ and then translated with $t$. Both $s$ and $t$ are learnable parameters. The transformation matrix is below:

\begin{equation}
	\begin{bmatrix}
		a & 0 \\
		0 & 1
	\end{bmatrix}
	\begin{bmatrix}
		1 & t \\
		0 & 1
	\end{bmatrix} = 
	\begin{bmatrix}
		a & a \cdot t \\
		0 & 1
	\end{bmatrix} 
\end{equation}

The initial values for $s$ and $t$ are from a normal distribution with mean zero and stdev of 0.15 and 0.1

\subsubsection{Gaussian}

In \eqref{eq:gauss_kernel} I use the exponent of a variable denoted $\sigma$ as a standard deviation because it will better match the dynamic range of the network's learning capabilities. A $\sigma$ parameter of 0 will correspond to a Gauss kernel with standard deviation of 1.

\begin{equation}
	\label{eq:gauss_kernel}
	f(x) = e^{\left( \frac{x - \mu}{e^\sigma}\right) ^2}
\end{equation}

The parameters $\sigma$ and $\mu$ are learnable and are initialized from a normal distribution with zero mean and a standard deviation of 0.1 and 0.01 * S/2 respectively, where S is the time length of the input signal (Figure \ref{fig:data_division}). Note the drop of the scaling coefficient. This has no impact on the learning capabilities and increase the computation speed by a small factor.

\subsubsection{LogNormal}

There is a basis kernel generated using \eqref{eq_logNormal}. This kernel is then shifted so the mode is at the 0 position ($x=e^{-1}$).

\begin{equation}
	\label{eq_logNormal}
	k(x) = \frac{1}{x} e^{-\frac{\left( \log{x} \right)^2}{2}}
\end{equation}

To generate the convolutional filter, the basis kernel $k$ is scaled and translated using an 1D affine transform with learnable parameters scale $s$ and $t$:
\begin{equation}
	\begin{bmatrix}
		a & a \cdot t \\
		0 & 1
	\end{bmatrix}
\end{equation}

where $a=e^{s}+3$ and $s$ and $t$ are drawn initially from a normal distribution with zero mean and stdev of 0.5 and 0.1.

\subsubsection{Gabor}
There are two steps in bandpass filtering because the filter has complex (real and imaginary) response. The first pass is a convolution like in previous filters and the second step is computing the magnitude and orientation from the filter responses.

The learnable parameters are $s$ and $\mu$ and the outputs of this module are $mag(t)$ and $ang(t)$. In equations \eqref{eq:gabor_filter} the $sig(t)$ is the one dimensional feature vector that will be filtered and $\circ$ is the convolution operation.

\begin{eqnarray}
	\omega&=&\text{sigmoid}(s)(0.5 - \frac{2}{S}) + \frac{2}{S}\nonumber\\
	\sigma&=&\omega\sqrt{\frac{\log 2}{\pi}} \frac{2^\text{bw}+1}{2^\text{bw}-1} \nonumber\\
	re(x)&=&e^{-\frac{1}{2}\left(\frac{x-\mu}{\sigma} \right)^2} \cos{\left(2\pi\omega (x-\mu)\right)}\nonumber\\
	\label{eq:gabor_filter}
	im(x)&=&e^{-\frac{1}{2}\left(\frac{x-\mu}{\sigma} \right)^2} \sin{\left(2\pi\omega (x-\mu)\right)}\\
	o_{re}&=&sig \circ re\nonumber\\
	o_{im}&=&sig \circ im\nonumber\\
	mag(t)&=&\sqrt{o_{re}(t)^2 + o_{im}(t)^2}\nonumber\\
	ang(t)&=&\arctan\left(\frac{o_{im}(t)}{o_{re}(t)}\right)\nonumber
\end{eqnarray}

The bandwidth $\text{bw}$ is set to 2.5 octaves. The learnable parameters are $s$ and $\mu$ initialized from a normal distribution with mean 2 for $s$ and zero for $\mu$ and standard deviation of 1 for $s$ and 0.2/S for $\mu$. In the formulas above, S is the time length of the input signal and $\text{sigmoid}(x) = 1/(1+e^{-x})$. A Gabor layer will double the feature count for a signal.

There are several options for bandpass filtering. I chose Gabor instead of LogGabor because Gabor filters have DC response \cite{vicas_curvilinear_2015}. That is, they are sensistive to absolute values in the signal and keeping these absolute values between different features is desired. However, other applications might request a filter with slightly different properties. It is important to note that the flexibility of differentiable programming would allow implementing such a filter focusing only on the forward step.

\subsubsection{Causal Convolution}

A system that is designed to predict some future samples and which has access to those samples can just otput those known instances. To avoid this, one must take extra care that for predicting a time step $t$, no future information is involved into the computation. One way to achieve this is through causal convolutions. These are regular convolutions but with a bit of extra padding in order for the input aperture of each output location not to extend beyound the temporal step for that output location. 

\begin{figure}[htbp]
	\begin{center}
		\begin{tikzpicture}[scale=0.6, x={(0.7cm,-0.5cm)}, y={(1cm,0cm)}, z={(0cm,1cm)}] 
			\begin{scope}[plane origin={(0, 0, 0)}, plane x={(1, 0, 0)}, plane y={(0, 1, 0)}, canvas is plane]  
				\fill [fill=green!15!white] (0,0) rectangle (1, 9);
				\fill [fill=green!35!white] (2,4) rectangle (3, 12);
				\draw [->, thick] (0, 12) -- (3.5, 12) node[below] {features};  
				\draw [->, thick] (0, 0) -- (0, 12.5) node [above, ] {time};   
				\draw [step=1] (0, 0) grid (3, 12);
			\end{scope}
			\begin{scope}[plane origin={(0, 8, 3)}, plane x={(1, 8, 3)}, plane y={(0, 9, 3)}, canvas is plane]  
				\fill [fill=green!15!white] (0,0) rectangle (1, 1);
				\fill [fill=green!35!white] (2,3) rectangle (3, 4);
				\draw [->, thick] (0, 4) -- (3.5, 4) node[below] {features};  
				\draw [->, thick] (0, 0) -- (0, 4.5) node [above, ] {time};   
				\draw [step=1] (0, 0) grid (3, 4);
			\end{scope}
			\draw [color=gray, very thin] (0.5, 0, 0) -- (0.5, 8, 3);
			\draw [color=gray, very thin] (0.5, 9, 0) -- (0.5, 9, 3);
			
			\draw [color=gray, very thin] (2.5, 4, 0) -- (2.5, 11, 3);
			\draw [color=gray, very thin] (2.5, 12, 0) -- (2.5, 12, 3);
			
		\end{tikzpicture}
		\caption{Causal convolution. No output cell can "see" beyond its current timestamp.}
		\label{fig:causal_convolution}
	\end{center}
\end{figure}
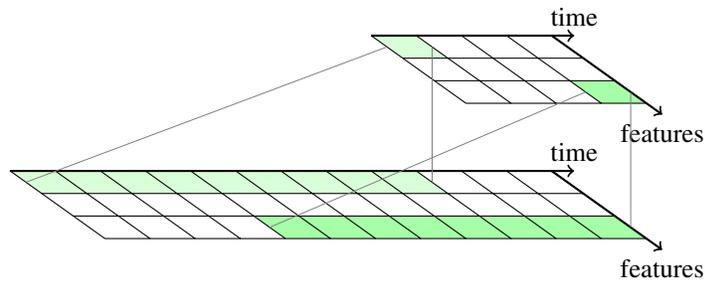

Figure \ref{fig:causal_convolution} shows how this causal convolution works.

\subsubsection{Feature aggregation}

The second major inductive bias is that once the time delay is elliminated, the output can be generated as a function of the inputs. The most general form of function approximator [from the neural network architectures] is the fully connected architecture (FCN). This is a stack of several linear layers with a non-linear activation between them.

Here, this activation is  LeakyReLu operation shown in Equation \eqref{eq:leakyrelu}. The last layer is not follwed by activation so the block can be directly used in regression problems.

\begin{equation}
	\text{LeakyReLu(x)} = \text{max}(x, 0) + 0.01\cdot\text{min}(x, 0)
	\label{eq:leakyrelu}
\end{equation}

The parameters of this block are:

\begin{itemize}
	\item The number of intermediate layers (0 means that there will be only one layer that processes the input)
	\item The number of features for the intermediate layers, as a multiplier of the input size
	\item The number of output features
\end{itemize}

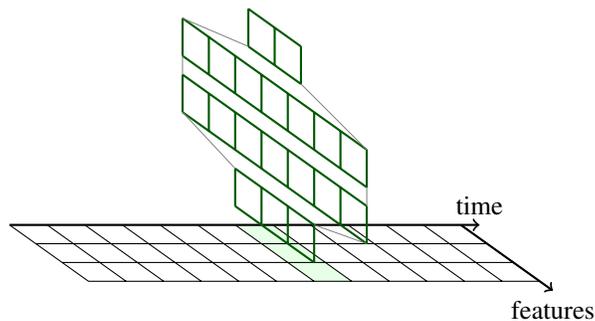
\begin{figure}[htbp]
	\begin{center}
		\begin{tikzpicture}[scale=0.5, x={(0.7cm,-0.5cm)}, y={(1cm,0cm)}, z={(0cm,1cm)}] 
			
			\begin{scope}[plane origin={(0, 0, 0)}, plane x={(1, 0, 0)}, plane y={(0, 1, 0)}, canvas is plane]  
				\fill [fill=green!10!white] (0,6) rectangle (3, 7);
				\draw [->, thick] (0, 12) -- (3.5, 12) node[below] {features};  
				\draw [->, thick] (0, 0) -- (0, 12.5) node [above, ] {time};   
				\draw [step=1] (0, 0) grid (3, 12);
			\end{scope}
			
			\begin{scope}[plane origin={(1, 5.3, 1)}, plane x={(2, 5.3, 1)}, plane y={(1, 5.3, 2)}, canvas is plane, 
				color=green!35!black]   %
				\draw [thick] (0, 0) grid (3, 1);
				\draw [shift={(0,0.5)}, thick] (-2, 1) grid (5, 2);
				\draw [thick] (-2, 3) grid (5, 4);
				\draw [shift={(0.5,0.5)}, thick] (0, 4) grid (2, 5);
				\draw [color=gray,very thin] (0, 1) -- (-2, 1.5);
				\draw [color=gray, very thin] (3, 1) -- (5, 1.5);
				
				\draw [color=gray, very thin] (-2, 2.5) -- (-2, 3);
				\draw [color=gray, very thin] (5, 2.5) -- (5, 3);
				
				\draw [color=gray, very thin] (-2, 4) -- (0.5, 4.5);
				\draw [color=gray, very thin] (5, 4) -- (2.5, 4.5);
				
			\end{scope}

		\end{tikzpicture}
		\caption{Feature aggregation. The input aperture of the fully connected network (drawn in dark green) spans accros features and does not have access to other values than the current timestamp.}
		\label{fig:featurewise_aggregation}
	\end{center}
\end{figure}

Note in Figure \ref{fig:featurewise_aggregation} how this FCN is applied to the features. All the inputs are aligned in time and each time position is processed with the same FCN network. Note that the input is not the original feature set but the delayed features.

\subsection{Architecture}
 
\begin{figure}[htbp]
	\begin{center}
		\begin{tikzpicture}[node distance = 0.2 and 0.7]
			\tikzstyle{every node}=[font=\small]
			\node (in1) [nn-data]  {$X_1$: (B, F, S)};
			\node (in2) [nn-data, right=of in1]  {$X_2$: (B, C, T)};
			\node (filter-low) [nn-layer, below=of in1]  {Filter Low: (B, n$\cdot$F, S)};
			\draw [->] (in1) -- (filter-low);
			\node (agg-low) [nn-layer, below=of filter-low] {Aggregator Low: (B, Fc, S)}; 
			\draw [->] (filter-low) -- (agg-low);
			\node (relu-1) [nn-layer, below=of agg-low] {LeReLu}; 
			\draw [->] (agg-low) -- (relu-1);
			\node (temporal-agg) [nn-layer, below=of relu-1] {Temporal Agg: (B, Fc, T)}; 
			\draw [->] (relu-1) -- (temporal-agg);
			\node (relu-2) [nn-layer, below=of temporal-agg] {LeReLu}; 						
			\draw [->] (temporal-agg) -- (relu-2);
			\node (concat) [nn-layer, below=of relu-2, xshift=35] {Concatenate: (B, Fc+C, T)};
			\draw [->] (relu-2) -- (concat.north-|relu-2);
			\draw [->] (in2) -- (concat.north-|in2);
			\node (filter-mid) [nn-layer, below=of concat] {Filter High: (B, n$\cdot$(Fc+C), T)};
			\draw [->] (concat) -- (filter-mid);
			\node (relu-3) [nn-layer, below=of filter-mid] {LeReLu}; 						
			\draw [->] (filter-mid) -- (relu-3);
			\node (agg-mid) [nn-layer, below=of relu-3] {Aggregator high: (B, Fy, T)};						 						
			\draw [->] (relu-3) -- (agg-mid);
			\node (out) [nn-data, below=of agg-mid] {Out: (B, Fy, T)};
			\draw [->] (agg-mid) -- (out);
		\end{tikzpicture}
		\caption{Overall \textit{Delay} architecture. \textbf{B} is batch size, \textbf{F} is the number of features in the known past ($X_1$); \textbf{S} is the time sequence length of the known past, \textbf{T} is the time sequence length of the future, \textbf{C} is the number of commands, \textbf{Fy} is the number of features in the target. \textbf{Fc} is the number of features after the first aggregation. It acts as a bottleneck. The parameter \textit{n} denotes how many filters per feature the Filter Block has.}
		\label{fig:overall_arhitecture}
	\end{center}
\end{figure}
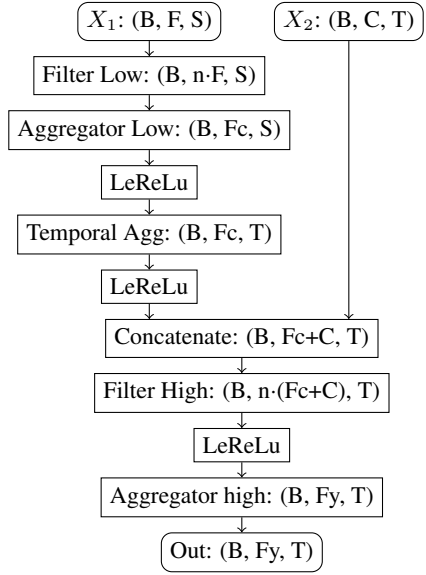

 There are many ways in which the building blocks can be mixed and matched to generate a functional neural network. In Figure \ref{fig:overall_arhitecture} is the \textit{Delay} architecture. 
 
The time delay of the features and the way they interact is resolved by coupling a filtering block with a feature aggregation block. These two blocks exploit the separability of the temporal influence versus feature interaction.

There are of course, assumptions made. One filtering layer is enough to "align" the features so they are relevant for deriving the output values. Only some aperture from these filtered features is relevant for the future values. The aperture is selected using temporal aggregator block. The future commands will impact the future evolution of the system of course, with some delay. There is another succesion of filterng and feature aggregation that acts upon the concatenation between the selected features and the future commands the system will receive.

This second stack of filters and aggregators has an activation between them. The intention was to have a learning process happening. As in any ReLu activation family, negative values would be filtered out and would not impact much the final feature aggregator. This activation is missing from the input filter-aggregator block because, there, negative values carry a meaninful physical information. First, one can have negative sensor information (temperature), second a high-pass filter regulary produces negative values. Placing a nonlinearity would force the BatchNorm component to learn a high bias, just to preserve these negative values for future use.

A note to the reader, this is not the first architecture that was proposed, several failed attempts are listed at the section \ref{chapter:attempts}. However, for this particular architecture, the only parameter that was changed from "original" attempt was the replacement of ReLu with LeakyRelu activation, between the building blocks. No other searching/tuning was performed except for the hyperparameters of each individual block.

\section{Results}

\subsection{Data}

The data source is identical to the one described in \cite{vicas_iccp2020}. The data was collected during a 3 year period and splitted in train and validation sets. Validation is in the future of training. There are several special validation subsets where a certain condition was observed. In Table \ref{tab:data_splits} are the specifications. Note that a bit more aggresive data curation leads to sligtly lower volumes than presented in previous work.
Here, the data preprocessing is done in a slightly different manner:

\begin{enumerate}
	\item Original sensor data has a sample ratio of one minute.
	\item Each sensor/data column has the missing values inputed with a linear interpolation if the gap between two consecutive values is less than 20 minutes
	\item Data is windowed with a certain stride (\textasciitilde{50 minutes}). Windows with missing values are rejected. In previous work, the missing data was inferred inside the window. This is the source of reduced data volumes in Table \ref{tab:data_splits}.
	\item To generate a sample datapoint I average three timestamps. A sample length of 160 will cover 8 hrs of sensor data.
	\item The sample is split into known past (\textbf{S} = 100 timestamps) and predicted future (\textbf{T} = 60 timestamps)
	\item \label{item:averaging} Features are grouped by their meaning and physical properties (eg one group consists of all temperature readings). The mean and standard deviation are computed and substracted accros each group. 
	\item \label{item:substract} 20\% from the known target values are averaged and substracted from \textit{all} temperature columns (including the future target). The percent is not that important but some small tests revealed that a lower value is better. Keeping only one sample resulted in sligthly unstable training for \textit{Delay} architectures.
\end{enumerate}

\begin{table}[htbp]
	\caption{Training and evaluation data. Certain validation subsamples met special conditions (shown in \textbf{Observations} column.) The \textit{train} set is included in \textit{train-large} and \textit{train-mild} is a subset of \textit{train}.}
	\begin{center}
		\begin{tabular}{|l|c|c|c|}
			\hline
			\textbf{Name}&\textbf{Period}&\textbf{Count}&\textbf{Observations} \\
			\hline
			\textit{train}&Winter Y1&4K&\makecell[lc]{Main train dataset}\\
			\hline
			\textit{train-large}&\makecell[lc]{Autumn Y1 -\\ Autumn Y2}&13K&\makecell[lc]{Extended dataset including \\spring and summer of Y2.}\\
			\hline
			\textit{train-mild}&Winter Y1&2.4K&\makecell[lc]{Days with outside temp.\\ \textless  -10\textdegree C were removed}\\
			\hline
			\hline
			\textit{validation}&Winter Y2&6.6K&Main validation set\\
			\hline
			\textit{val-cold}&Winter Y2&0.36K&\makecell[lc]{Only days with outside temp.\\ \textless \hspace{1pt} -10\textdegree C are included.}\\
			\hline
			\textit{val-quiet}&Winter Y2&0.2K&\makecell[lc]{Few days without human\\ presence.}\\
			\hline
		\end{tabular}
		\label{tab:data_splits}
	\end{center}
\end{table}

The \ref{item:averaging}'th step is very important for the \textit{Delay} networks. The \textit{Attention} based architectures are not that affected by changing the data preprocessing. Current implementation allows one to group the features according to their properties (eg. temperature sensors, voltage readings, angles, distances) and each group will be normalized independently of other groups. The values inside a group are normalized with the same statistic. This is a compromise between keeping the dynamic range of the original signal and conditioning the data for neural networks.

While all the architectures in \cite{vicas_iccp2020} had the BatchNorm as the first transformation, for \textit{Delay} type architectures this was not the case. Also, during architecture exploration, a BatchNorm before the delay lines hurt the numerical stability of the learning process.

Note that for all architecutres, the \ref{item:substract}'th step will increase the training speed and stability. The network does not have to learn the actual values of the target, just the trend in the near future. This is true of course only for certain targets, where a degree of continuity and a small 1st order derivative is expected (eg. it varies slowly). For our problem, it means that a network predicting zero, will in fact, predict that there will be no change in the temperature with respect to last seen values. Which is perfectly reasonable assumption for some systems.

\subsection{Evaluation methodology}

The following filters are used for low and high level feature expansion: Identity, Gauss, LogGauss, Affine, Gabor. These filters can work in \textit{one-filter-per-feature} or  \textit{one-filter-per-cell} setup.
Temporal aggregation layer uses previous filters but only in \textit{one-filter-per-cell} setup and the CausalConvolution "filter".

After few iterations, the Gabor filter was elliminated because of numerical instabilities. A quick look revealed that the issue is in the derivative of the magnitude computation. No more effort was placed in solving this issue, as it is unclear why it surfaces only when Gabor layer is integrated in deeper architectures. 

The training loss and reported loss is Mean Absolute Error:

\begin{equation}
	\text{MAE}=\frac{1}{N}\sum_{0}^{N}|target - prediction|
\end{equation}

In regression problems another measure is frequent: Root Mean Squared Loss (RMSE). RMSE tend to induce a network that learns the larger variations in the signal. Here, I am more interested in learning the small variations and less worried about large misclassifications. In practice, these large errors could be handled by a safeguard system.

For reporting purposes (eg charts) the predictions are scaled back with the same statistics used for preprocessing and the target is the actual target temperature in Celsius. To show the results, I will use \textit{boxplots}. Each box encloses the 25-75 percentile. Whiskers are at 10-90 percentile and the dots are outliers. The horizontal line inside the box is the median value of the plotted data.

\subsection{Results}

The baseline for these experiments are the \textit{Attention} networks developed in previous work \cite{vicas_iccp2020}. The proposed architecture here, has a lot of hyperparameters and their effect and interactions cannot be inferred. I moved forward using an iterative approach, at each iteration certain experiments were performed and conclusions were drawn. Next iteration was based on previous conclusions and findings. 

All the details regarding the training loop (learning rate, schedulers, early stopping, etc) can be found in the available source code and they will not be presented here. Mostly, they follow known trends in literature \cite{bag_tricks_2019, glorot10a}.

The first batch of experiments involved performing a random grid search on the type of filters, number of filters, number of layers and kernel widths in temporal aggregators, and other architecture parameters while keeping the dataset parameters constant.

Using a logistic regression I determined the most important parameters and did a drilldown of the results. Unfortunately few clear conclusions could be drawn. The type of filters is important. \textit{One-filter-per-feature} gives better performance than \textit{one-filter-per-cell} mode. I selected few architectures for the next step and added three Attention based networks as baselines.

\begin{itemize}
	\item Filter low: Affine and LogGauss, 2-8 filters, 
	\item Aggregator low: 2-8 layers, and expansion ratio 0.5 - 1.5
	\item Bottleneck: 8-16 features
	\item Temporal aggregator: Causal, Affine
	\item Filter high: Affine, Gauss, 2-8 filters
	\item Aggregator high: 2-8 layers and expansion ratio 0.5 - 1.5
\end{itemize}

Again, a random search was performed.

From this second run, I chose the following two architectures:

First architecture, named \textit{D\_AffAffGau} is:

\begin{itemize}
	\item Filter low: Affine 4 filters
	\item Aggregator low: 2 layers, and expansion ratio 1
	\item Bottleneck: 8 features
	\item Temporal aggregator: Affine
	\item Filter high: Gauss 8 filters
	\item Aggregator high: 2 layers and expansion ratio 1
	\item 12.3K weights for 5 feature dataset and 17.4K for 13 feature dataset
\end{itemize}

Second architecture, named \textit{D\_LogAffGau} is:

\begin{itemize}
	\item Filter low: LogGauss 4 filters
	\item Aggregator low: 8 layers, and expansion ratio 1
	\item Bottleneck: 8 features
	\item Temporal aggregator: Affine
	\item Filter high: Gauss 8 filters
	\item Aggregator high: 2 layers and expansion ratio 1
	\item 14.9K weights for 5 feature dataset and 33.9K for 13 feature dataset
\end{itemize}

For the second architecture there are more layers in the first agrregator and different low filter count. \textit{LogAffGau} gave the best performance as a group, although the best representative was on the 2'nd position overall. The best performing instance was from \textit{AffAffGau}.

For this batch of experiments, two \textit{Attention} based networks were selected too: 

A small network, with few layers and features named \textit{Att\_A1}

\begin{itemize}
	\item Hidden size: 8
	\item Number of LSTM layers: 4
	\item Dropout: 0.35
	\item 7K weights for 5 feature dataset and 7.2K for 13 feature dataset
\end{itemize}

A beefy network, with more layers and features named \textit{Att\_A3}

\begin{itemize}
	\item Hidden size: 512
	\item Number of LSTM layers: 4
	\item Dropout: 0.35
	\item 12.2M weights for 5 feature dataset and 12.3M for 13 feature dataset
\end{itemize}

The "SOTA" network in the \cite{vicas_iccp2020} with 1 layer and 256 hidden units was called in previous experiments as \textit{Att\_A2} and its performance was between the much smaller \textit{Att\_A1} and beefier \textit{Att\_A3}. For brevity its results were excluded from further computation and plotting.

For this experiment I trained and evaluated with both the reduced feature set (5 features) and extended set (13 features). The only parameter that was allowed to take random values was the random seed used to initialize the weights and randomize the training data/batches. The results presented below are plotted from this batch of experiments. Also, except where specified, the results are on \textit{train} dataset, with the evaluation on \textit{validation} dataset, using 5 features.

\begin{figure}[htbp]
	\centerline{\includegraphics[scale=0.9]{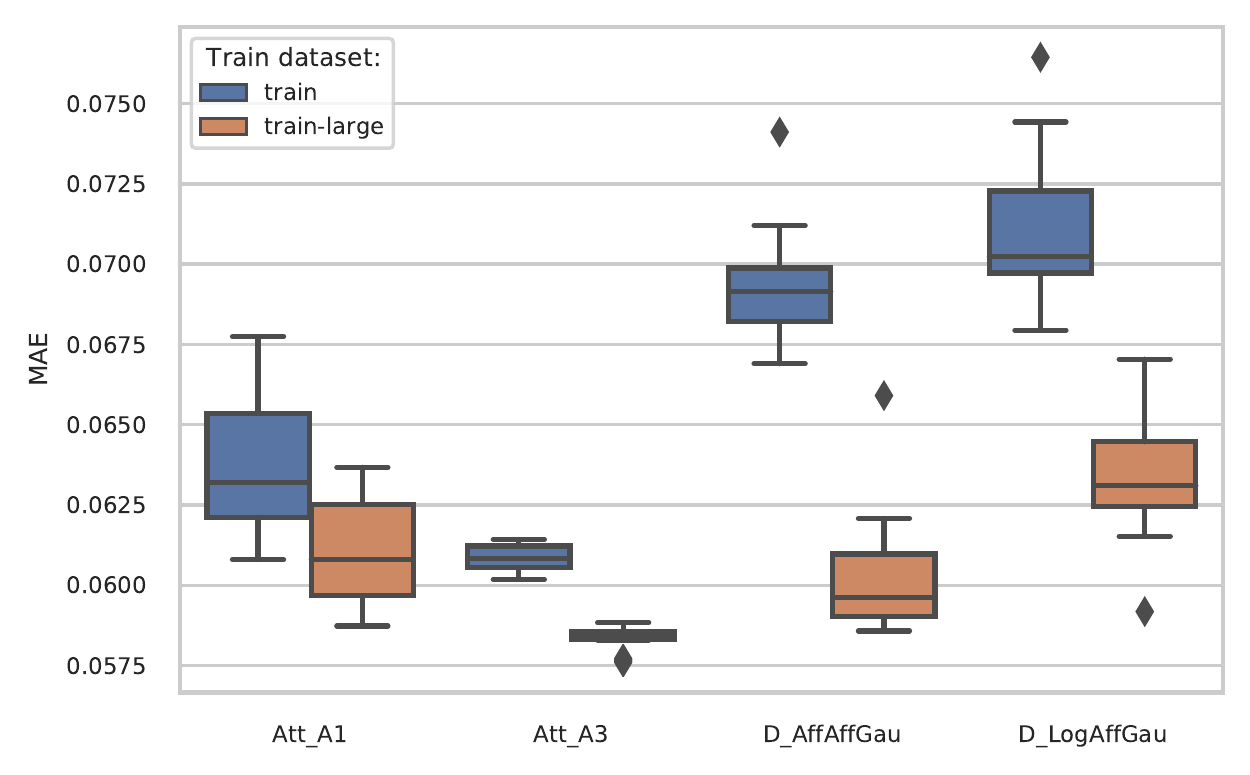}}
	\caption{Overall performance on \textit{validation} dataset for selected architectures, using 5 features.}
	\label{fig:perf_each_arch_valid}
\end{figure}

\begin{figure}[htbp]
	\centerline{\includegraphics[scale=0.9]{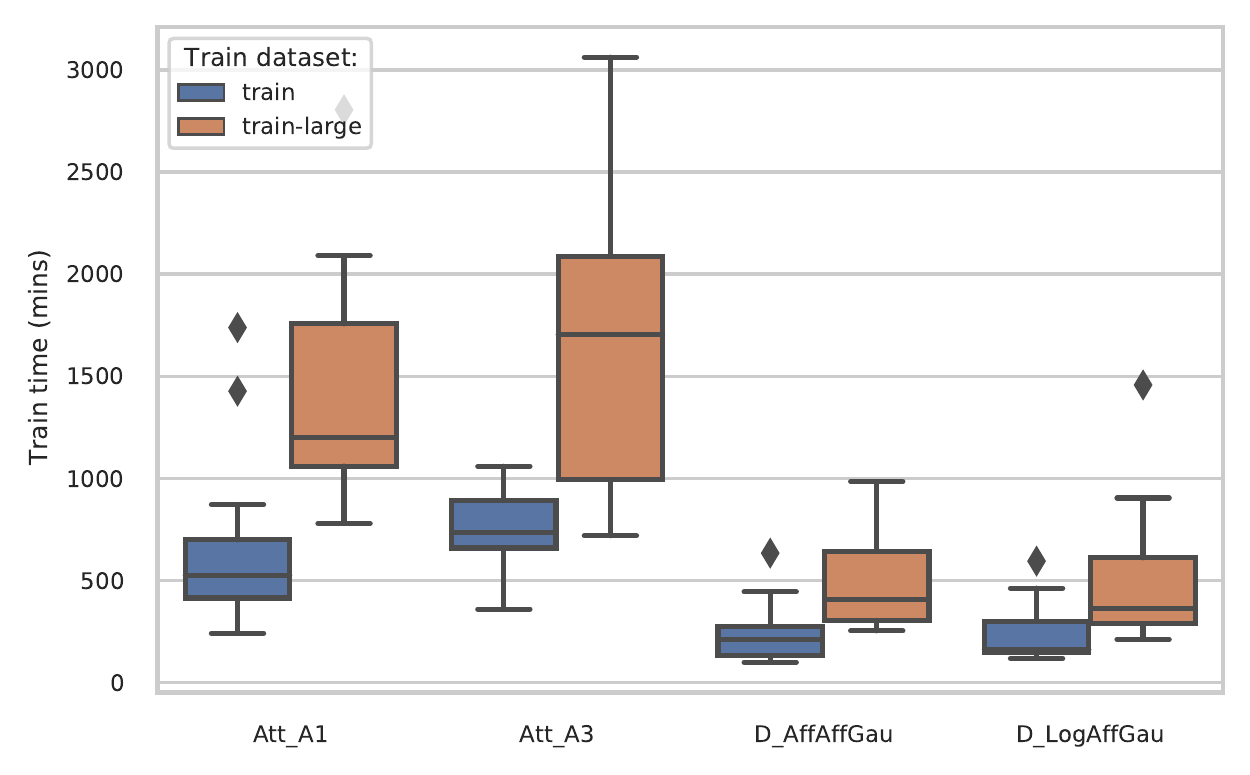}}
	\caption{Training times on selected architectures.}
	\label{fig:train_time_each}
\end{figure}

\begin{figure}[htbp]
	\centerline{\includegraphics[scale=0.9]{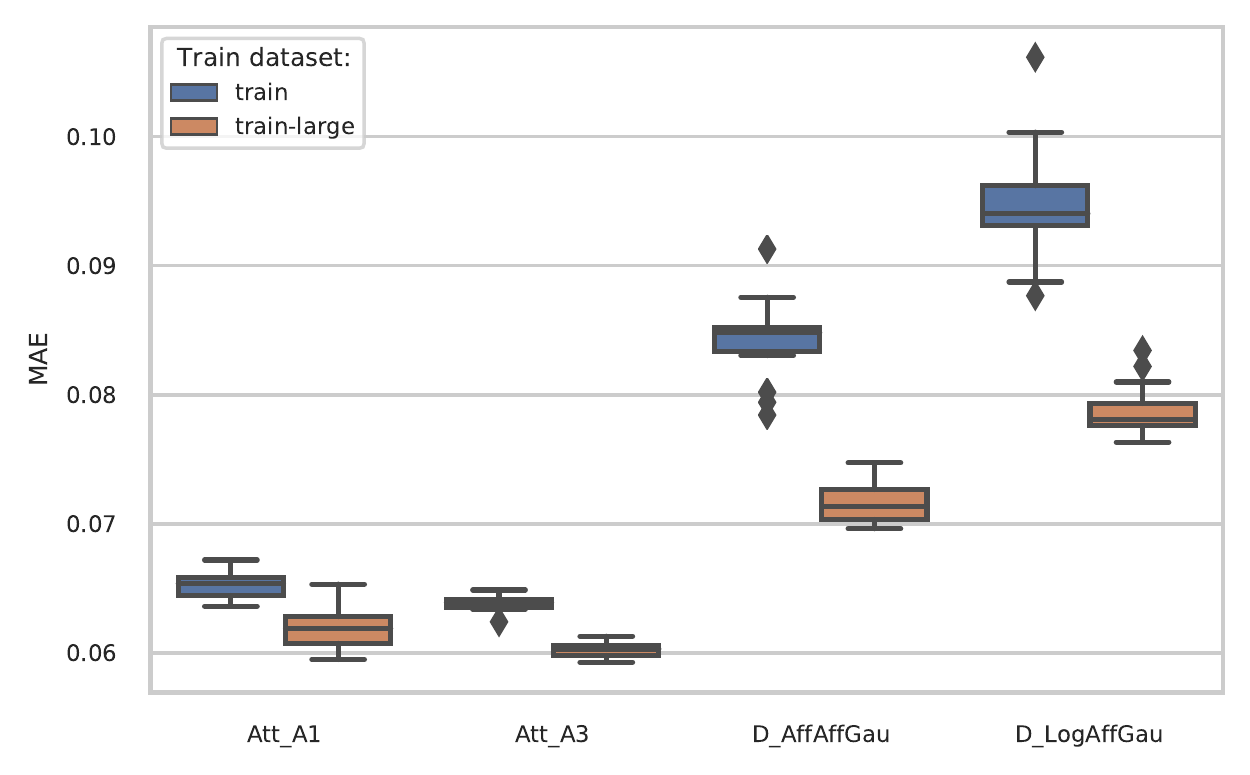}}
	\caption{Overall performance on \textit{validation} dataset for selected architectures using 13 features.}
	\label{fig:perf_each_arch_valid_13}
\end{figure}

\begin{figure}[htbp]
	\centerline{\includegraphics[scale=0.9]{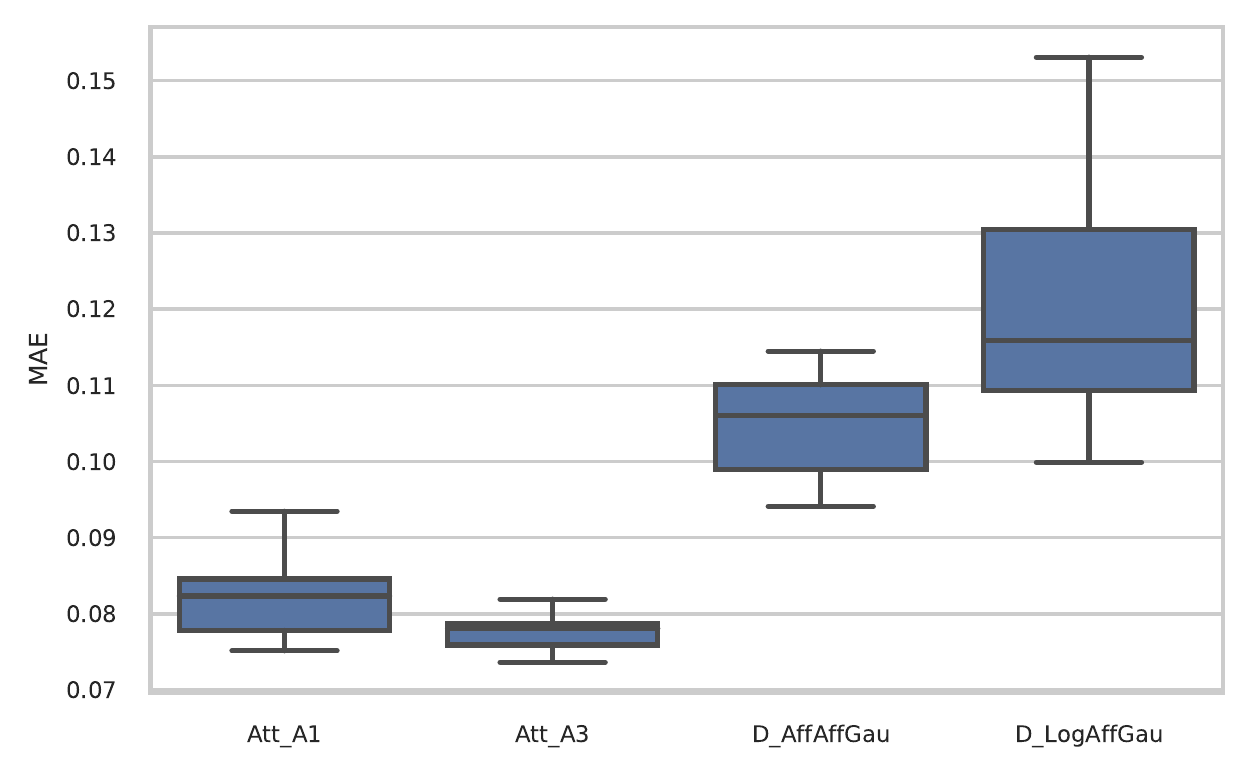}}
	\caption{Performance when the training set does not contain hard winter data and the validation set contains only hard winter data.}
	\label{fig:perf_hard_winter}
\end{figure}

\begin{figure}[htbp]
	\centerline{\includegraphics[scale=0.9]{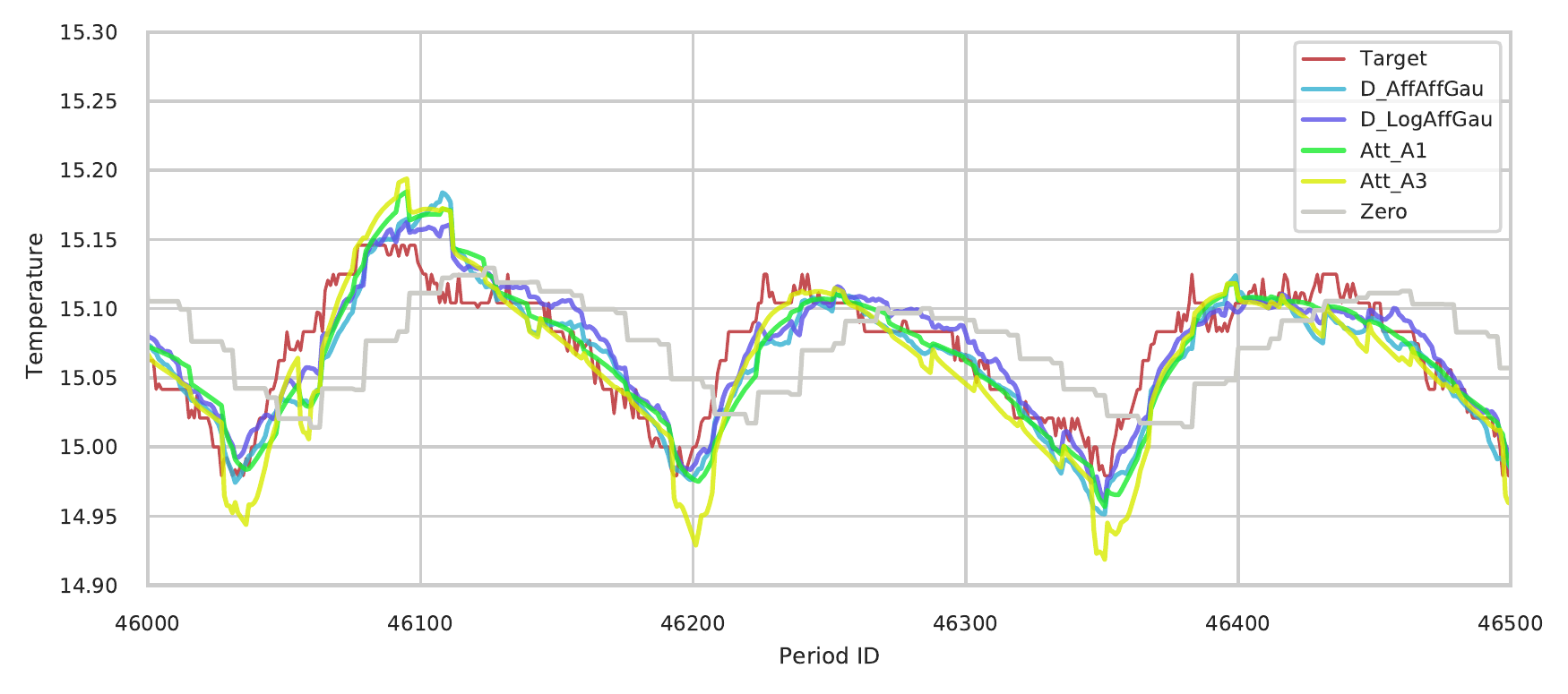}}
	\caption{Sample of data, for a quiet (no human presence) period. One x value is approx 3 minutes. Fine gray line is the \textit{Zero} predictor.}
	\label{fig:sample_quiet}
\end{figure}

\begin{figure}[htbp]
	\centerline{\includegraphics[scale=0.9]{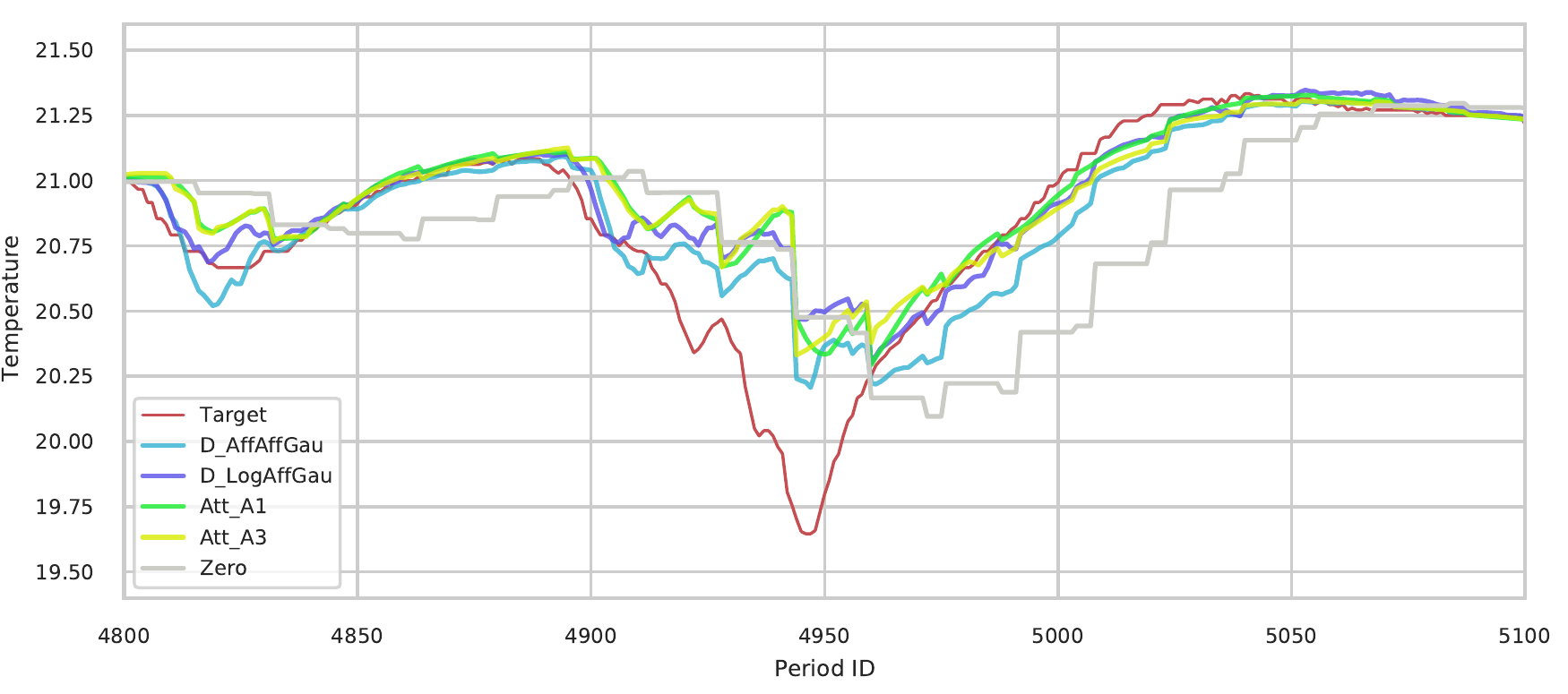}}
	\caption{Sample of data, for a venting event. One x value is approx 3 minutes.  Fine gray line is the \textit{Zero} predictor.}
	\label{fig:sample_venting}
\end{figure}

\begin{figure}[htbp]
	\centerline{\includegraphics[scale=0.9]{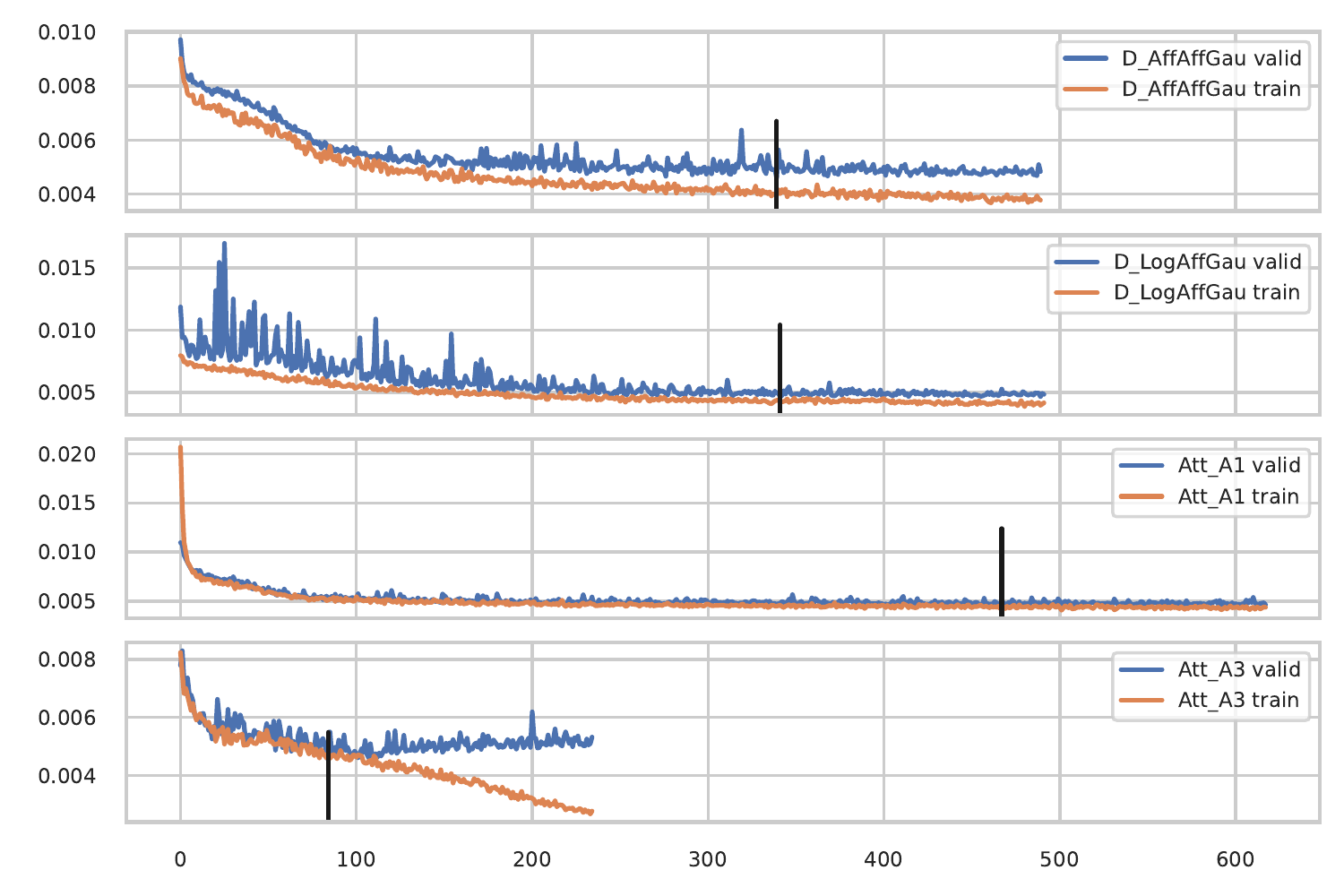}}
	\caption{The training curves for the best network samples for each architecture. Note that the x axis is the epoch number. The black line represents the position where the best validation loss was observed and is the checkpoint that is saved for later use. The Att networks, despite needing less epochs, take significantly more wall time to train.}
	\label{fig:training_curves}
\end{figure}

\begin{figure}[htbp]
	\centerline{\includegraphics[scale=0.9]{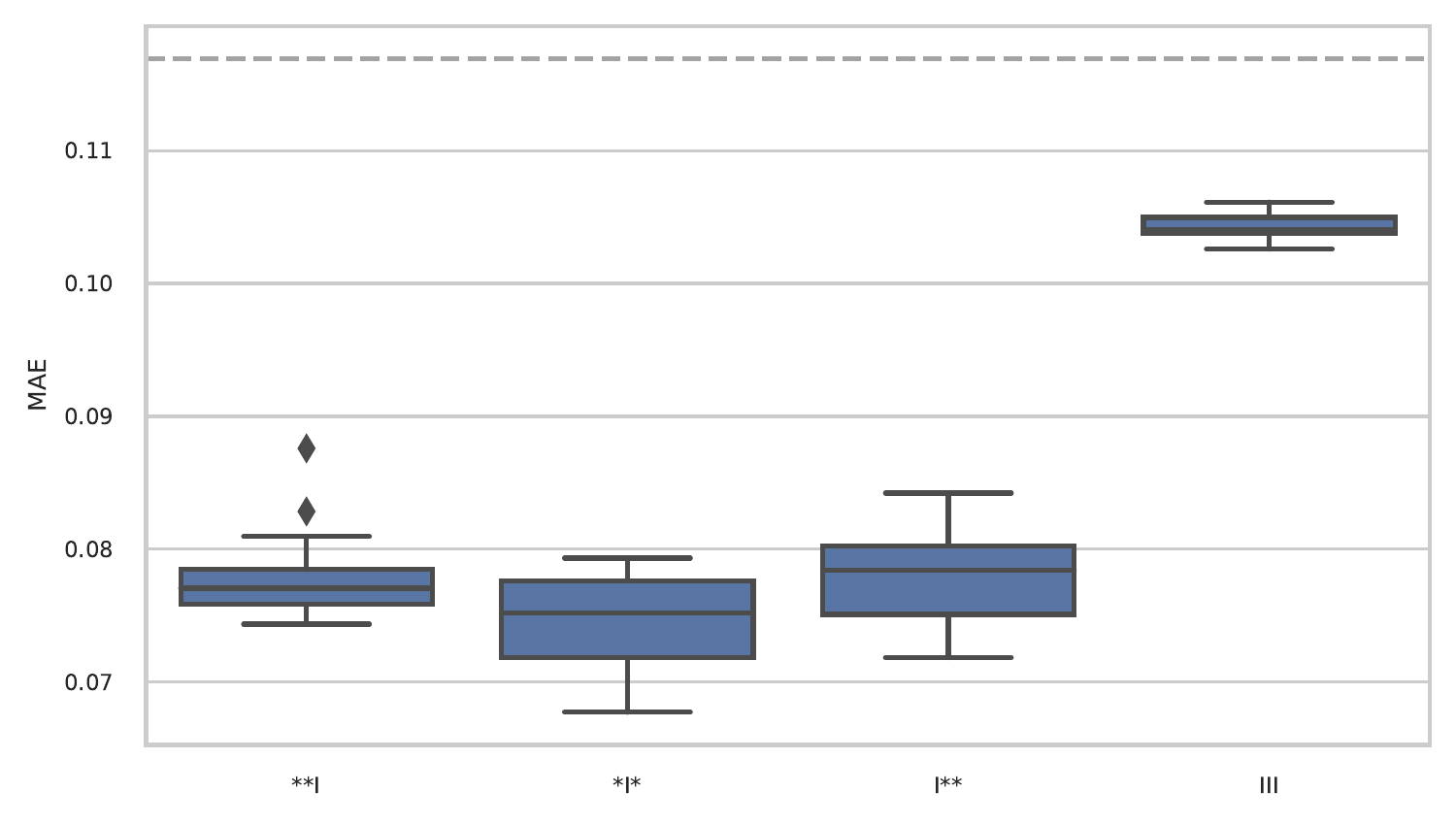}}
	\caption{Performance when one of the filtering layers was replaced by Identity. Eg: *I* means that the temporal aggregation layer was replaced by Identity while the rest of the filtering layers (Expander low and high) could take any value except Identity. The dashed line represents the MAE loss when the raw output is always zero. That is, the network just outputs an average of the last known target values}
	\label{fig:ablation_identity}
\end{figure}

In Figure \ref{fig:perf_each_arch_valid} is the overall situation of the architectures. The training times are shown in Figure \ref{fig:train_time_each}

From a numerical point of view the \textit{Attention} based networks are the winners here, but the \textit{Delay} based networks are not lagging far behind. Looking at the best results, obtained on \textit{train-large} dataset, there is a 2.7\% drop in median MAE between \textit{D\_AffAffGau} with 0.06 and \textit{Att\_A3} with 0.0584 median MAE.  If the data volume is decreased (\textit{train}), the \textit{D\_AffAffGau} has a loss of 0.0691 and \textit{Att\_A3} 0.0607. That is, a drop of 12.1\% for the median loss.  Interestingly, the \textit{Att\_A3}  does not benefit much from increased data volumes.

Examining the training time, it reveals the advantage of architecture parallelization: The \textit{Delay} architectures train five to ten times faster than \textit{Attention} counterparts. Comparing the network sizes of approx 12M weigths for \textit{Att\_A3} and 17K for \textit{D\_AffAffGau} reveal that the efforts of searching for other architectures were successful.

Figure \ref{fig:perf_each_arch_valid_13} shows that more features (eg 13 instead of 5) are not helpful but the \textit{Delay} architecture can benefit more from increased number of samples. Both types of architectures take a penalty from having more features. Maybe additional features don't bring anything useful and only increase the network capacity?

Figure \ref{fig:perf_hard_winter} shows how the networks respond to out-of-train data. The training data was \textit{train-mild} and validation was done using \textit{val-cold}. Notably, the results are more differentiated. The \textit{Attention} is better with approx 25\% but is it because of the reduced data volumes? Are the \textit{Attention} networks using the data in a more efficent manner? Or the \textit{Attention} architecture really learns the system in a better way?

The best network samples from this last experiment (Figure \ref{fig:perf_each_arch_valid}) will be used for the rest of the results presented in this subsection.

In practice one would infer a new prediction each time a new data point is available as input. One way to aggregate this new prediction with existing predictions is with an exponential moving average window, where each past network prediction has exponentially lower contribution. Here, the sampling is done at 16 time periods (so each result is an exponential average of ~10 inference results) and each period in the plots is approx 3 minutes. A trivial predictor will output the mean observed temperature value, because one doesn't expect sudden jumps in temperature from one sample to another. Such a predictor, called \textit{Zero} was added to the results. Exponential aggregation was performed for it, too.

Before moving to the graphs, there is one important note to make. Best sample from \textit{Att\_A3} has a loss of 0.0576 and the best sample from the \textit{D\_LogAffGau} has a loss of 0.0591, within 2.5\% of each other. That is two orders of magnitude lower than the assumed measurement device error level.

In Figure \ref{fig:sample_quiet} and \ref{fig:sample_venting} I show several interesting cases from validation data. First sample (Figure \ref{fig:sample_quiet}) looks at a quiet period, where, for several days there was no human activity. There are several warming/cooling cycles, all networks perform virtually identical with the exception of the Zero network that has a non zero phase. 

Figure \ref{fig:sample_venting} shows a venting event. The windows are opened to the outside air for a certain period of time. At the beginning of this event, all networks are misspredicting the future temperature by a large margin. Their performance is close to the \textit{Zero} network. However, after a short time (enogh information about the venting event is in the known past) all networs start to correctly predict future temperature (except, of course, for \textit{Zero} "network").

Examining the training curbes, \textit{Att\_A1} has the lowest overfitting capability and \textit{D\_LogAffGau} has a bit of unstable training. The training curves presented in Figure \ref{fig:training_curves} are for the best samples of the last experiment.

In short, a good architecture is \textit{D\_AffAffGau} architecture with fairly low capacity. The wall training times are better than for \textit{Attention}, the capacity of learning from new data is better than for \textit{Att\_A1} and only within 3\% distance from the "best" network, \textit{Att\_A3} both trained on larger dataset.

\section{Ablation Studies}

Here, I try to answer the meta-questions: What makes the network/method work? What are the critical parts? The easiest way to do this is to "disable" certain parts of the pipeline and note the degradation of the results.

Data handling surfaced as the first critical element. If not done in the specific way presented here, will yield performance degradation and training instabilities (NaNs, loss divergence). 

Coupled with data preprocessing and making sure that the data ranges remain stable, the focus moves to the first BatchNorm in the network. This BN layer is after the Filtering convolutions. Removing or changing the post filtering BatchNorm layer had an impact on the results. Removing this layer or placing a simple featurewise multiplication/translation learnable layer gave slightly worse results. This is interesting because BatchNorm layer basically ensures that any discrepancy between input features (in terms of dynamic range) is removed.

Removing all (or some) of the filtering layers has detrimental impact on the results. In Figure \ref{fig:ablation_identity} I ran several experiments where one of the filtering networs are replaced by Identity layer. The rest of the layers can have one of the previously presented filters. The number of filters and other parameters were kept as for \textit{D\_AffAffGau} architecture.

Notaby, having no translation capable layers, and relying simply on spatio/temporal separability of the network yields a performance close to a \textit{Zero} outputing network (dotted line in Figure \ref{fig:ablation_identity}). 

Having filtering layers improve the results by a large factor. Among the three possible positions, the lower filtering block has the biggest importance. Surprinsigly, having an Identity in the temporal aggregator has lesser importance and some networks (at least in these experiments) had samples reaching the performance ranges of \textit{D\_LogAffGau} architecture.

The \textit{Delay} based architectures benefit more from higher data volume. However, (very interesting) if this volume is obtained using a data augmentation technique, there is no performance gain. This was tested on two "champion" networks, \textit{D\_AffAffGau} and \textit{D\_LogAffGau} by sampling the training data at a smaller stride (0.01 instead of 0.1 as in all previous experiments). Basically, the data volume was increased tenfold. The gain in \textit{loss} was minimal. The difference between the \textit{train} and \textit{train-large} is the inclusion of more time intervals so more data diversity. As a result, not much effort is worth putting in inventive data augmentation techniques but rather in better exploiting the available data (eg hard case mining, activation decay, etc). Unfortunately the wall time for \textit{Att\_A1} and \textit{Att\_A3} was too big for this setup and this analysis was not performed for them.

\section{Discussions and Conclusion}
\subsection{Attempts that did not work}
\label{chapter:attempts}
Like in any research project, from the idea to the realization there were several unsuccesfull attempts. I list them here, maybe some will be worth re-visiting in the future.

\textbf{BatchNorm} layer plays an important role in DL in general. It helps with network convergence and dynamic range discrepancies. Initial assumption was that the relative differences between temperatures matter. The data preprocessing routine takes this assumption into account. Removing (or replacing) BatchNorm layer with something that retains these relative differences \textit{after} the first Filter block yielded poorer results. Reciprocal is also true. Adding a BatchNorm or any other feature-wise scaler \textit{before} first Filter block rendered the network untrainable.

\textbf{Future features prediction.} One architecture choice that made sense "on paper" was to use some part of the network to predict the future feature values (ex. outside disturbances), concatenate them to the known features and then, use the delay layers plus an aggregator to predict the unkwnon target. I used a loss with two components, one for the future features and one for the target. The network learned something useful when the first loss was close to zero. Moreover, if the future feature prediction component had a non trivial capacity (a fully connected stack or a LSTM/GRU) all the learning happened inside this component. Maybe for this particular problem the actual inertia is so large that the future values of our features are not relevant, or there are some other forces involved.

\textbf{Custom derivatives} for filtering layers. All the filters are derivable with respect to their parameters and their input. For some layers I computed these derivatives analytically and integrated into the DL framework. Unfortunately the computation speed was lower than using auto-differentiation. This idea will be revisited in the future because, for Gabor filters (or at least for computing the magnitude), an approximation for the derivatives of that layer will be needed.

Using \textbf{sub-networks} to derive the filter parameters. In the architecture presented here, the filter parameters are learned as any other parameter. But this is not the only option. These parameters could be computed by another part of the network and that part of the network is where the learning will happen. Unfortunately, except for some fully connected layers, there is no \textit{a priori} knowledge that could be used to derive the architecture of the sub-network. Having so many weights in the FCN part led to overfitting and worse results. The same thing happened when using \textit{one-filter-per-cell} mode. All except in the temporal aggregation layer (where \textit{one-filter-per-feature} doesn't make sense) the \textit{one-filter-per-cell} mode yielded bad results. Probably, in a problem with far more data, making the translation/scale parameters depend on input could be achieved.

\subsection{Discussions and future directions}

Qualitative analysis (Figure \ref{fig:sample_quiet}, \ref{fig:sample_venting}) shows that the outputs are highly correlated so future efforts should be placed in gathering relevant data from the environment rather than future tuning of the architectures. No network, no matter how large capacity it has (eg \textit{Att\_A3} with 12M parameters) can "predict" if a person just opened a window, based solely on sensors presented here. Also, from a physical point of view, opening the windows drastically change the system to the point that it has no resemblance to the "original" one. The networks now have to learn two subsystems without having hints on when the transition between these two states occur.

The strong corellation between the network outputs suggests that the results shown in Figure \ref{fig:perf_hard_winter} are probably because the \textit{Attention} models are more data efficent and not because the network learns in a better way, the system.

Looking at Figure \ref{fig:perf_each_arch_valid} and Figure \ref{fig:perf_hard_winter} one can note that increased data volumes correlates with better performance. The rate is higher for the \textit{Delay} based architectures than for the \textit{Attention}. Evidence from the \textit{scaling laws} \cite{Jones2021} in ML shows that one can extrapolate the data volume/performance without performing cumbersome experiments. As a result, maybe the existing \textit{Attention} based architectures, especially the \textit{Att\_A3} should be skipped from future experiments and the computation power should be redirected to exploiting ways to replace the FCN component in \textit{Delay} networks.

The \textit{D\_AffAffGau} architecture will be put into production, hopefully for the next winter season. It has the advantage of being quick to train (so, for given compute capacity the network can be further tuned) and can benefit from new data points. \textit{Attention} based networks don't have these advantages while only sporting minor prediction performance gains.

There is no need for this particular problem to dive into reduced-precision computations (eg 16, 8 bit) but, for other problems these avenues m be exploited.

The FCN subnetwork that handles the feature interaction is responsible for most of the large network weight count and overfitting. Probably it can also explain the low network performance in the presence of new features. Having one extra feature means that there are eight new input features to the FCN stack. Being a FCN, the number of connections per layer grows with $O(n^2)$ so $16n+64$ new weights per layer per new added feature. This is where a residual like connection might be useful. Maybe for different types of problems the practitioner might already know some interaction patterns between features and these interactions could be directly coded into the aggregator subnetwork, letting the differentiable programming paradigm derive the actual parameters. More thought must be put here to see how to derive a computational unit that can be connected into a residual setup. Would it make sense? Adding a shifted/transformed version of the input to itself? Would it resemble some sort of derivative of the original signal? And while on vanishing gradients issue, the Gabor filters must be also adapted to work for deeper networks. There are other tricks that help with learning performance, like hard case mining and activation decay, tricks that will be exploited in the future.

Another avenue in replacing the FCN layers, left for a future work, is to create a delay/aggregate block that generates one or few features at the output. The input of this block, consisting of many features, delays the signal then "selects" the important features that will be aggregated. The selection can be performed using a (tempered) SoftMax or directly, using a linear layer with strong L1 weight decay. Many of such 1 feature blocks can be stacked horizontally (in parallel) and vertically (concatenate their outputs and forward it to another batch of blocks). This stacking is similar to how Transformers are constructed. Hints that this heavily constrained feature aggregation might be a good strategy emerged very late in the research, while preparing the public software+data package.

\subsection{Conclusion}

In present paper, the "intuitions" about the physical phenomenon were translated into powerful inductive biases that drove the neural network architecture. I managed to exploit the time/feature separability intuition and self supervised framework to devise a network that is faster and yields similar performance results as the classical (and very powerful) reccurent encoder-decoder with attention architecture. When the data volumes are increased, the \textit{Delay} architecture median loss drops from \textasciitilde 15\% behind \textit{Attention}, to only \textasciitilde 3\%. Best individual model instances from each architecture were basically indistinguishable, from a qualitative standpoint.

The data preprocessing is very important. In hindsight, this shouldn't be a surprise, in any DL field, a bad data curation leads to bad results (images must be standardized, in NLP the dictionary must be curated, token positions must be added to Transformer architectures, etc.).

The reader should keep in mind that recurrent architectures and Attention mechanisms are rather stable and mature while these filtering based architectures received far less attention. It is possible to gain future non trivial gains by investing more energy in this avenue.

\bibliographystyle{abbrv}
\bibliography{bibauto2}






\end{document}